%% file: main.tex
\documentclass{article}

\PassOptionsToPackage{numbers, compress}{natbib}

\usepackage[preprint]{neurips_2022}

\usepackage[utf8]{inputenc} 
\usepackage[T1]{fontenc}    
\usepackage{hyperref}       
\usepackage{url}            
\usepackage{booktabs}       
\usepackage{amsfonts}       
\usepackage{nicefrac}       
\usepackage{microtype}      
\usepackage{xcolor}         
\usepackage[algo2e]{algorithm2e}
\usepackage{algorithm}
\usepackage{algpseudocode}
\usepackage{caption}
\usepackage{subcaption}
\usepackage{floatpag}

\title{Remember and Forget Experience Replay for Multi-Agent Reinforcement Learning}

\author{%
  Pascal Weber, Daniel Wälchli, Mustafa Zeqiri, and Petros Koumoutsakos\thanks{Corresponding Author petros@seas.harvard.edu.}\\
  Computational Science and Engineering Laboratory\\
  ETH Zürich, Switzerland \\
  and \\
  John A. Paulson School of Engineering and Applied Sciences \\
  Harvard University, USA
}

\usepackage{amsmath,amssymb,amsthm}
\usepackage{hyperref}
\usepackage{cleveref}
\usepackage{color} 
\usepackage{bbold}
\usepackage{graphicx}
\usepackage{algorithm}
\usepackage{array}

\definecolor{modelInd}{RGB}{78, 115, 174}
\definecolor{modelIndMA}{RGB}{219, 132, 87}
\definecolor{modelCoop}{RGB}{88, 167, 106}
\definecolor{modelCoopMA}{RGB}{194, 79, 84}

\definecolor{brown}{RGB}{147,122,99}
\definecolor{pink}{RGB}{217, 141, 194}
\definecolor{purple}{RGB}{129, 116, 177}

\definecolor{additions}{RGB}{78, 115, 174}

\input{macros}

\begin{document}

\maketitle

\begin{abstract}
We present  the extension of the Remember and Forget for Experience Replay (ReF-ER) algorithm  to Multi-Agent Reinforcement Learning (MARL).
ReF-ER was shown to outperform state of the art algorithms for continuous control in problems ranging from the OpenAI Gym to complex fluid flows. 
In MARL, the dependencies between the agents are included in the state-value estimator and the environment dynamics are modeled via the importance weights used by ReF-ER. In  collaborative environments, we find the best performance when the value is estimated using individual rewards and we ignore the effects of other actions on the transition map. 
We benchmark the performance of ReF-ER MARL on the Stanford Intelligent Systems Laboratory (SISL) environments. We find that employing a single feed-forward neural network for the policy and the value function in ReF-ER MARL, outperforms state of the art algorithms that rely  on complex neural network architectures.
\end{abstract}

\section{Introduction}
State of the art deep reinforcement learning (RL) algorithms approximate the optimal value function and policy using deep neural networks. This approach has been showcased in the playing of Atari games~\cite{Mnih2015}, board games like Sh$\bar{\text{o}}$gi, Chess, and Go~\cite{Silver2017}. More recently Multi-Agent Reinforcement Learning (MARL) has been applied to  multiplayer games such as poker~\cite{Brown2019} and prominent computer games such as Dota 2 and StarCraftII~\cite{OpenAI2019,Vinyals2019}. 
Tasks that require multiple agents to collaborate often rely on a  generalization of single agent RL  algorithms and  employ complex neural network architectures for the value estimation. Deep MARL presents several algorithmic challenges. Learning individual policies is hard in environments with multiple learning agents that can be non-stationary. Moreover, the inclusion of collaborative behaviour by an averaging of the rewards yields a credit assignment problem, hindering the reinforcement of beneficial behaviour. 
Finally, most games are restricted to a discrete action space. More recently the extension of MARL to scientific computing and complex systems (scientific multi-agent reinforcement learning (SciMARL)~\cite{Novati2021,Bae2022}) has showcase the importance of control in continuous, high-dimensional action spaces. To the best of our knowledge, generalizations of well-established algorithms for continuous action RL is limited~\cite{Gupta2017,terry2020revisiting}.

We address these challenges by revisiting the relationships and interactions between multiple agents. We follow the centralized training with decentralized execution paradigm~\cite{Oliehoek2018,Kraemer2016} (CTDE). Agents learn using the experiences from all their peers, while at execution time, the learned policy is used to make decisions based on a single agent's state information. 
The inclusion of the observations from all agents addresses the non-stationary issues, and the adoption of ReF-ER handles effectively far-policy experiences~\cite{Novati2019}. Furthermore, the learning rule can be modified to systematically examine the credit assignment problem. Lastly, ReF-ER allows to model the interaction strength between agents via the importance weight.

\section{Multi-Agent Reinforcement Learning}\label{sec:marl}

\begin{figure*}
	\centering
	\includegraphics[width=\linewidth,trim=75 225 75 230, clip]{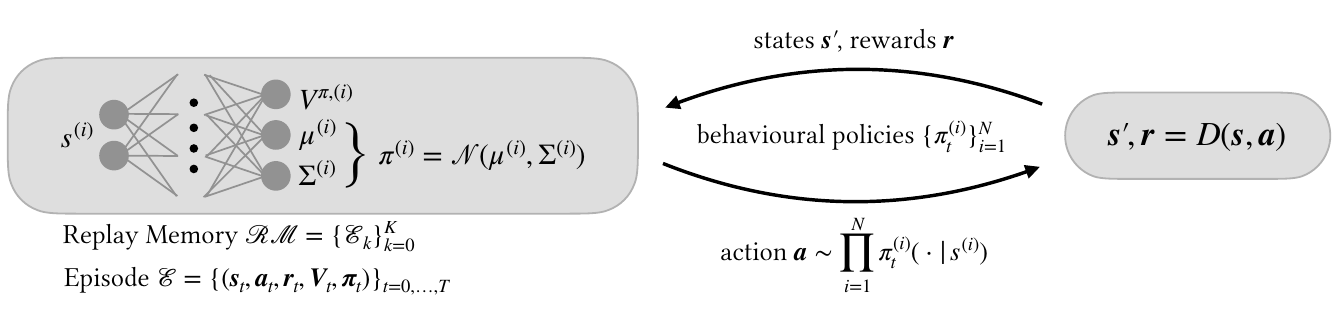}
	\caption{Schematic of the MARL loop.}\label{fig:MARL-loop}
\end{figure*}

Multi-Agent Reinforcement Learning (MARL) aims to find the optimal policy in a Multi-Agent Markov Decision Process (MAMDP). A MAMDP is a tuple $(\stS, \acS, \vreward, N, D)$ consisting of the states $\vstate\in\stS^N$, actions $\vaction\in\acS^N$, and rewards $\vreward\in\mathbb{R}^N$ observed by the $N\in\mathbb{N}_+$ agents in the environment.
Here we consider a homogeneous setting, where all agents have the same state space, action space, and reward function.
The agents' individual states, actions, and rewards are denoted by $\st^{(i)}\in\stS$, $\ac^{(i)}\in\acS$, and $\re^{(i)}\in\mathbb{R}$ for $i=1,\dots,N$. 
The transition map is given by $\vreward,\vstate'=D(\vstate,\vaction)$, i.e. it depends on the states and actions from all agents.
This is contrary to environments in which agents take actions sequentially and the state is updated after each action. 
In MAMDP we specify the initial state distribution $p(\vstate)$ and the stochastic policy ${\pi}(\vaction|\vstate)$. The later is the probability distribution over the actions of the agents given the states. We assume that the policy factorizes as 
\begin{equation}\label{eq:policy}
{\pi}(\boldsymbol{a}|\boldsymbol{s})=\prod_{i=1}^N\pi(a^{(i)}|{s}^{(i)})\PERIOD
\end{equation}
This allows for \textit{decentralized execution}, where the agent samples an action solely based on its state.
The vector-valued state-action-value $\boldsymbol{Q}$ function and state-value $\boldsymbol{V}$ are defined as
\begin{equation}
\boldsymbol{Q}^{\pi}(\vstate,\vaction)=\mathbb{E}_{\pi}\left[\sum\limits_{t=0}^\infty \gamma^t \boldsymbol{\re}_t|\vstate_0=\vstate, \vaction_0=\vaction\right]\COMMA\quad \boldsymbol{V}^{\pi}(\vstate)=\mathbb{E}_{\pi}\left[\boldsymbol{Q}^{\pi}(\vstate,\vaction)\right]\PERIOD
\end{equation}
where $\gamma\in [0,1)$ is the discount factor and $\boldsymbol{r}_t=\boldsymbol{r}(\vstate_t,\vaction_t)$. Assuming equal importance for all agents yields the scalar state-value that is expressed in terms of the individual state value functions $V^{\pi}(\sti)$
\begin{equation}\label{eq:goal}
P\left[\boldsymbol{V}^{\pi}(\vstate)\right]= \frac{1}{N}\sum\limits_{i=1}^N V^{\pi}(\sti)\PERIOD
\end{equation}
The optimal policy $\pi^\star$ in the MAMDP maximizes the average of the state-values for all agents 
\begin{equation}
{\pi}^\star=\arg\max\limits_\pi\frac{1}{N}\sum\limits_{i=1}^N V^{\pi}(\sti)\COMMA\; \forall \boldsymbol{s}\in{\cal S}^N\PERIOD
\end{equation}
In MARL the agents interact with the environment as depicted in \cref{fig:MARL-loop}. At timestep $t$, the agents take actions $\vaction_t$ based on the states $\vstate_{t}$. Thereby the environment transitions into new states $\vstate_{t+1}$ and returns rewards $\vreward_t$. An \textit{experience}, consists of states, actions, and rewards $(\vstate_t,\vaction_t,\vreward_t)$. Upon termination of the episode ${\cal E}_k=\{(\vstate_t,\vaction_t,\vreward_t)\}_{t=1}^{T_k}$ the experiences are stored in the replay memory ${\cal RM}=\{{\cal E}_k\}_{k=1}^K$. In the following $d$ denotes the distribution of the experiences in the replay memory.

We extend V-RACER with ReF-ER~\cite{Novati2019} for MARL, training a neural network that approximates the state-value and the policy parameters. Using the experiences in the replay memory, we train the neural network with Adam~\cite{Kingma2015}. In the following, $\pi_{\boldsymbol{\omega}}(\ac|\st)$ and $V^{\pi}_{\boldsymbol{\vartheta}}\left(\st\right)$ denote the policy and state-value function with the respective weights $\boldsymbol{\omega}$ and $\boldsymbol{\vartheta}$. Note, that the parameters $\boldsymbol{\omega}$ and $\boldsymbol{\vartheta}$ only differ in the output layer.

\subsection{Value Learning}\label{sec:value learing}
The relationships between the agents is included in the learning process by introducing a scalarization function \mbox{$f:\mathbb{R}^N\to \mathbb{R}$} in the on-policy returns estimator Retrace~\cite{Munos2016}
\begin{equation}
\begin{split}
\hat{Q}_{t,f}^{\mathrm{ret}}= f(\boldsymbol{\re}_t) + \gamma V_f^\pi(\vstate_t)+ \gamma \bar{\rho}_t(\hat{Q}_{t+1,f}^{\mathrm{ret}} - Q_f^\pi(\vstate_t, \vaction_t))\PERIOD
\end{split}
\end{equation}
V-RACER approximates $Q_f^\pi(\vstate_t, \vaction_t)\approx V_f^\pi(\vstate_t)$ and hence the V-trace estimator~\cite{Wang2016} becomes
\begin{equation} \label{eq: V-trace estimator}
\hat{V}_{t,f}^{\mathrm{tbc}}=V_f^\pi(\vstate_t)+\bar{\rho}_{t}\left[f(\boldsymbol{\re})_t+\gamma \hat{V}_{t+1,f}^{\mathrm{tbc}}-V_f^\pi(\vstate_t)\right]\COMMA
\end{equation}
and it can be related to the on-policy returns estimator Retrace via
\begin{equation}
\hat{Q}_{t,f}^{\mathrm{ret}}=f(\boldsymbol{\re})_t+ \gamma \hat{V}_{t+1,f}^{\mathrm{tbc}}\PERIOD
\end{equation}
The truncated importance weight $\bar{\rho}_t$ will be defined in the following section. It balances the impact from off-policy data on the current estimator of the state-value. 

For the function $f$ we distinguish two cases. The first case, which we refer to as \textit{individual}, assumes that the value estimate depends solely on the reward observed by agent $i$
\begin{equation}\label{eq:individual Vtbc}
\begin{split}
f_{\text{individual}}^{(i)}(\boldsymbol{r})=r^{(i)}\COMMA\quad 
V_{\text{individual},(i)}^{\pi}(\vstate)=V^{\pi}(\sti)\PERIOD
\end{split}
\end{equation}
In the second case, referred to as \textit{cooperative}, the reward and the state-values are averaged
\begin{equation}\label{eq:cooporative Vtbc}
\begin{split}
f_{\text{cooporative}}(\boldsymbol{r})= \frac{1}{N}\sum\limits_{i=1}^N r^{(i)}\COMMA\quad
V_{\text{cooporative}}^\pi(\vstate)=\frac{1}{N}\sum\limits_{i=1}^NV^{\pi}(\sti)\PERIOD
\end{split}
\end{equation}
The weights $\boldsymbol{\vartheta}$ of the neural network $V^{\pi}_{\boldsymbol{\vartheta}}\left(\st\right)$ are updated to minimize the loss
\begin{equation}
\label{eq:loss}
{\cal L}(\boldsymbol{\vartheta})=\mathbb{E}_{ d}\left[\frac{1}{N}\sum\limits_{i=1}^N\left(V_{\boldsymbol{\vartheta}}^{\pi}(\st_t^{(i)})-\hat{V}_{t,f}^{\mathrm{tbc}}\right)^2\right].
\end{equation}
Note, that both variants reduce to the original loss proposed in~\cite{Novati2019} when assuming a single agent.

\subsection{Policy Gradient}\label{sec:policy gradient}
Given the definition of $\hat{Q}_{t,f}^{\mathrm{ret}}$, we learn the weights $\boldsymbol{\omega}$ of the policy $\pi_{\boldsymbol{\omega}}(\st|\ac)$ by maximizing the expected advantage
\begin{equation}
\label{eq:advantage}
\mathcal{L}(\boldsymbol{\omega})=\mathbb{E}_{d}\left[\frac{1}{N}\sum_{i=1}^N\rho_t(\boldsymbol{\omega})\left(\hat{Q}_{t,f}^{\mathrm{ret}}-V^\pi(\st_t^{(i)})\right)\right].
\end{equation}
Taking the gradient with respect to the weights $\boldsymbol{\omega}$ yields the off-policy gradient~\cite{Degris2012}.
The importance weights $\rho_t(\boldsymbol{\omega})$ reflect the assumed dynamics of the environment.

Adopting the \textit{full dynamics} $\vreward^{t+1}, \vstate^{t+1}=D(\vstate^t,\vaction^t)$ and the factorization of the policy from~\cref{eq:policy} yields the importance weight
\begin{equation}\label{eq:dependent iw}
\rho_t^N(\boldsymbol{\omega})=\prod\limits_{i=1}^N \frac{\pi_{\boldsymbol{\omega}}(\ac_t^{(i)}| \st_t^{(i)})}{\pi(\ac_t^{(i)}| \st_t^{(i)})}\PERIOD
\end{equation}
Here the denominator denotes the policy that was used when $s^{(i)}$ was sampled. When using this importance weight the value estimate and the policy update depend on the probability of the action of all agents.

In cases where interactions between the agents are negligible, the value estimate and the policy update should not be influenced by the probabilities of the other agents. This is achieved by using the importance weight
\begin{equation}\label{eq:no dependency iw}
\rho_t(\boldsymbol{\omega})=\frac{\pi_{\boldsymbol{\omega}}(a_t^{(i)}| s_t^{(i)})}{\pi(a_t^{(i)}| s_t^{(i)})}\PERIOD
\end{equation}
We denote it as the \textit{local dynamics model}.


\subsection{Remember and Forget for Experience Replay}\label{sec:refer}
In order to update the policy, a mini-batch of experiences is sampled from the replay memory. For each of these experiences a gradient $\hat{\boldsymbol{g}}(\boldsymbol{\omega})$ is computed. ReF-ER avoids negative impact from off-policy data by classifying experiences based on the importance weight ${\rho}$. The criterion $\frac{1}{ c_{\max }}<\rho< c_{\max }$ depends on the cut-off $ c_{\max}$ that is annealed during training. The gradient is then regularized via
\begin{equation}
\begin{split}
\hat{\boldsymbol{g}}^{\text {R}}(\boldsymbol{\omega})=
\left\{\begin{aligned}
\beta \hat{\boldsymbol{g}}(\boldsymbol{\omega})-(1-\beta) \hat{\boldsymbol{g}}^{\mathrm{KL}}(\boldsymbol{\omega}) &\COMMA\;  \text {on-policy}\\
-(1-\beta) \hat{\boldsymbol{g}}^{\mathrm{KL}}(\boldsymbol{\omega}) &\COMMA\; \text {else.}
\end{aligned}\right.    
\end{split}
\end{equation}
The factor $\beta$ is adjusted according to the fraction of off-policy samples $n_{\text{off-policy}}$ in the replay memory
\begin{equation}\label{eq:beta update}
\beta \leftarrow \begin{cases}(1-\eta) \beta & \text { if } n_{\text {off-policy}}>n^\star \\ (1-\eta) \beta+\eta, & \text { otherwise.}\end{cases}
\end{equation}
Here $\eta$ denotes the learning rate used by Adam. This allows maintaining a target fraction $n^\star$ of off-policy experiences in the replay memory. The regularizer is the gradient of the Kullback-Leibler divergence between the current and the past policy~\cite{Schulman2015}
\begin{equation}
\hat{\boldsymbol{g}}^{\mathrm{KL}}(\boldsymbol{\omega})=\nabla_{\boldsymbol{\omega}}D_{\mathrm{KL}}\left(\pi_{k} \| \pi_{\boldsymbol{\omega}}\right)\PERIOD
\end{equation}
For details we refer to the publication of the ReF-ER algorithm~\cite{Novati2019}.

\section{Related Works}

In the earliest MARL studies  \cite{tan1993multi, Littman1994} the value-learning problem is solved via Q-learning with  each agent treating its peers as part of the environment.
Recent work in cooperative MARL, \citet{Sunehag2017} proposed  value-decomposition networks (VDN). The joint Q-function is modeled as the sum of individual Q-functions, which solely rely on the observation and action of a single agent. The individual Q-functions are trained by back-propagating the gradients from the Q-learning rule using the joint reward.
Thereafter QMIX was developed~\cite{Rashid2018} which uses deep recurrent Q-networks~\cite{Hausknecht2015} to represent individual state-action-value functions. The output of the Q-networks is combined with a mixing network to estimate a joint state-action-value-function. The weights of the mixing network are chosen such that the maximisation with respect to the actions of the agents yields the same optimum, independent whether the operation is applied on the mixed network or on the individual networks.
\citet{Son2019} propose QTRAN, which removes some of the structural constraints imposed by VDN and QMIX.
Along this,~\citet{Wen2020} proposed SMIX($\lambda$), where they benefit of the QMIX architecture combined with the novel variant of the off-policy algorithm SARSA($\lambda$)~\cite{Sutton2018} in order to learn the state-action value function.
\citet{Foerster2018b} proposed LOLA, which anticipates the policy change of the other agents by including a one-step look-ahead. This allows to maximize the returns, while considering that in the next update the agents will adjust their behaviour.
For continuous problems, \citet{Lowe2017} extended DDPG~\cite{Lillicrap2016} for the MARL setting. They learn a  centralized critic from which they train a deterministic policy for each agent. 
\citet{Gupta2017} compare the performance of policy gradient, temporal-difference error, and actor-critic methods in cooperative multi-agent systems. They  introduced the Stanford Intelligent Systems Laboratory (SISL) environments that are used as benchmarks in the present work. They evaluated their methods using different training schemes: centralized, concurrent and parameter sharing. In the first scheme they learn and execute in a centralized manner, which suffers from the curse of dimensionality. In the concurrent scheme they train independent policies for each agent and in parameter sharing they learn a single policy where the experiences from all agents are used during training.
This study was extended by \citet{terry2020revisiting}, where the impact of parameter sharing was evaluated on eleven modern single-agent RL algorithms. To the best of our knowledge, this study achieved the highest reported results on the SISL environments, and therefore we use it as the state of the art against which we compare our results.
\citet{Foerster2018a} proposed counterfactual multi agent policy gradients (COMA), which learns a centralized critic that is conditioned on the joint action, while each agent’s policy conditions on its observations. They address the credit assignment problem via a counterfactual baseline that compares the global reward to the obtained reward when that agent’s action is replaced with a default action. This mechanism allows to deduce the contribution of the single agent to the joint reward. However, this typically requires additional simulations to compute the reward using the default action. In order to facilitate this, they propose to learn an additional critic representation that allows an efficient computation of the baseline.
\citet{Christianos2020} proposes SEAC, which is an actor-critic algorithm that shares the experiences among all agents. The gradients for the actor and the critic have additional terms from the shared experiences, which are multiplied by an importance weight. By training multiple policies they can increase the exploration rate and by sharing experiences they facilitate learning in approximately equal rates.
FACMAC~\cite{Peng2020} learns a centralized but factored critic and gradient estimator. They have individual critic networks for each agent and combine them into a centralized critic using a non-linear mixing network similar as in QMIX~\cite{Rashid2018}. The difference is that they do not impose any constraints on the critic. They use a gradient similar to MADDPG~\cite{Lowe2017}, but instead of optimizing each agents actions separately, they resample actions for all agents with the current policy before calculating the gradient.

\section{SISL Environments}

\begin{figure*}
	\centering
	\includegraphics[width=\linewidth,trim=120 235 115 235, clip]{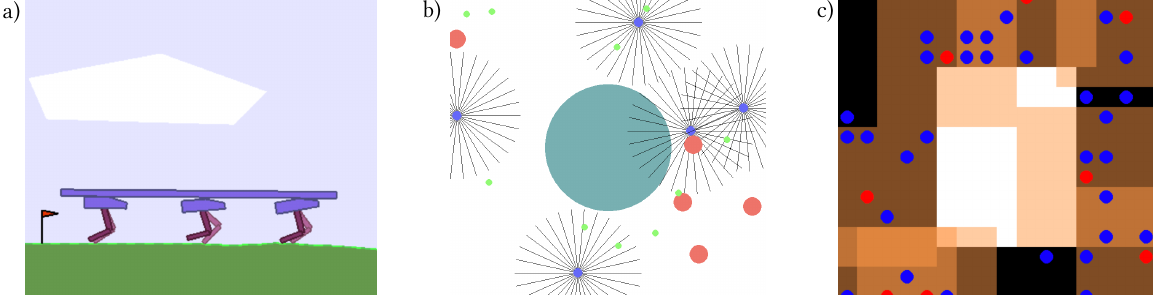}
	\caption{SISL environments: (a) Multiwalker, (b) Waterworld, and (c) Pursuit.}
	\label{fig:environments}
\end{figure*}

We  benchmark the performance of our algorithms in the three cooperative Stanford Intelligent Systems Laboratory Environments (SISL) implemented in PettingZoo~\cite{Gupta2017,terry2020pettingzoo} (see \cref{fig:environments}).

In \textit{Multiwalker} a package is located on top of three bipedal robots. 
The reward for each walker depends on the distance the package has traveled plus 130 times the change in the walker’s position. All agents receive -100 reward if any walker or the package falls. The walker that falls is further penalized by -10 reward. 
Each walker exerts force on two joints in their two legs, giving a continuous action space represented as a four element vector ${\cal A}_{\text{multiwalker}}\subseteq\mathbb{R}^4$. 
The state consists of 31 real values ${\cal S}_{\text{multiwalker}}\subseteq\mathbb{R}^{31}$ consisting of simulated noisy linear data about the environment and information about neighboring walkers. 
The environment ends after 500 steps or if the package or a walker falls.

In \textit{Waterworld} five agents attempt to consume food while avoiding poison. There are ten moving poison targets, which have a radius of 0.75 times the radius of the agent. Furthermore there are five moving food targets with radius two times the size of the agent radius. In the center of the domain is a solid object.
An agent obtains a shaping reward of $0.01$ for touching food. The food can be consumed if two agents touch it simultaneously, in which case both participating agents obtain 10 reward. On the other hand, touching a poison target and consuming it gives -1 reward. 
After consumption both food and poison targets randomly reappear at another location in the environment. 
The agents have a continuous action space represented by two elements ${\cal A}_{\text{waterworld}}\subseteq\mathbb{R}^2$, which corresponds to horizontal and vertical thrust.
In order to penalize unnecessary movement the agents obtain a negative reward based on the absolute value of the applied thrust. 
Each agents state results from 30 range-limited sensors, depicted by the black lines, which detect neighboring entities and result in a 242 element vector ${\cal S}_{\text{waterworld}}\subseteq\mathbb{R}^{242}$. The environment terminates after 500 steps.

In \textit{Pursuit} there are 30 blue evaders and eight red pursuer agents in a $16 \times 16$ grid with an obstacle in the center, shown in white.
Every time the pursuers fully surround an evader, each agent receives a reward of 5 and the evader is removed from the environment. In order to facilitate training, pursuers receive a shaping reward of 0.01 every time they touch an evader. 
The pursuers have a discrete action space, consisting of directions up, down, left, right or stay ${\cal A}_{\text{pursuit}}=\{\uparrow,\downarrow,\leftarrow, \rightarrow, \circ\}$. Each pursuer observes a 7 $\times$ 7 grid centered around itself, depicted by the orange boxes surrounding the red pursuer agents. For each of the gridpoints it receives three signals, the first signal indicates a wall, the second signal indicates the number of allies and the third signal indicates the number of opponents. Thus the observation space can be represented by ${\cal S}_{\text{pursuit}}\subseteq\mathbb{Z}^{147}$. The environment terminates after 500 steps or if all evaders are captured.

\subsection{Evaluation}

\begin{figure}
	\centering
	\includegraphics[width=\linewidth,trim=30 160 30 160, clip]{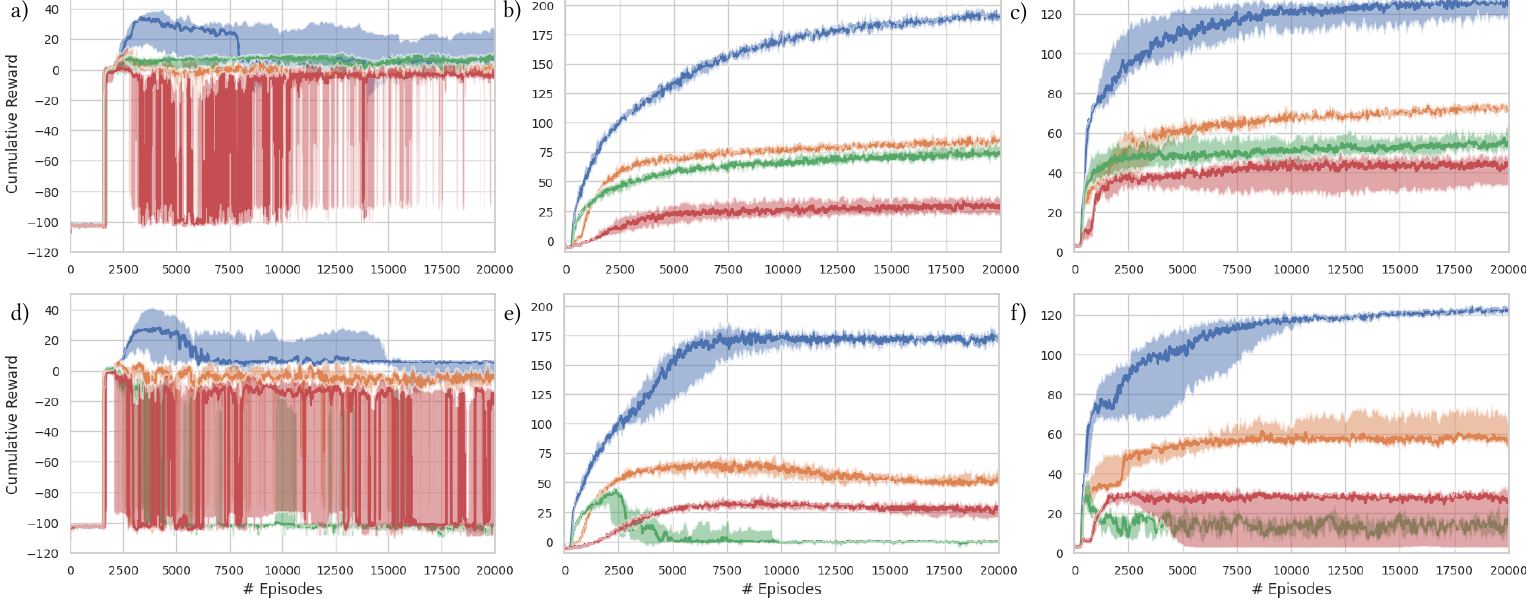}
	\caption{Median cumulative reward of 5 runs and 100 episodes, and averaged over all agents. The shading shows the 95\% confidence interval. The results in the first row (a-c) are for a single shared policy, the second row (d-f) for one individual policy per agent. The first column (a,d) shows Multiwalker, the second (b,e) Waterworld, and the third (c,f) Pursuit. The lines correspond to LDI~(\textcolor{modelInd}{\rule[0.5ex]{1.5em}{1.5pt}}), FDI~(\textcolor{modelIndMA}{\rule[0.5ex]{1.5em}{1.5pt}}), LDCo~(\textcolor{modelCoop}{\rule[0.5ex]{1.5em}{1.5pt}}), and FDCo~(\textcolor{modelCoopMA}{\rule[0.5ex]{1.5em}{1.5pt}}).}
	\label{fig:SISL results}
\end{figure}

We benchmark the algorithms using five runs with 20\,000 episodes each. For each run we compute the moving median of the cumulative reward averaged over all agents. The window size of the moving median is 100 episodes. In \cref{fig:SISL results} we plot the median of the medians and the 95\% confidence interval of the five runs. 
Throughout this study, we apply the default parameters suggested in~\citet{Novati2019}. We set the discount factor $\gamma = 0.995$, we use a replay memory of size $2^{18} = 262\,144$ with initial random exploration with a randomly initialized neural network for $2^{17} = 131\,072$ experiences. The learning rate of the Adam optimizer is initialized to $\eta=0.0001$ and the batchsize $B = 256$. For the approximation of the value function and the policy we use a neural network with two hidden layers of width 128. We initialize the ReF-ER factor $\beta = 0.3$, the target threshold $n^\star= 0.1$ and cut-off $c_{\max} = 4$. In contrast to the original V-RACER, we use the clipped normal distribution for continuous action domains and introduce a discrete variant of V-RACER (see Appendix). The source code is available on \url{https://github.com/will-be-provided} and we provide a pseudo code in the appendix. The SISL experiments where executed on a single CPU with 12 cores. The required training-times were approximately 48 hours per run. In total we performed 2x3x4x5=120 runs.

We distinguish the algorithms \cref{eq:dependent iw} that calculate the importance weights as V-RACER \textit{full dynamics individual-FDI} and \textit{full dynamics cooperative-FDCo}, depending on the value estimator (\cref{eq:individual Vtbc} and \cref{eq:cooporative Vtbc}). Accordingly, we name the algorithms V-RACER \textit{local dynamics individual-LDI} and \textit{local dynamics cooperative-LDCo}, if we apply \cref{eq:no dependency iw} to calculate the importance weight.
Finally, we distinguish training a single shared policy for all agents and one policy per agent. 
We sample a mini-batch for all variants and use the observations from all agents to train the neural networks using \cref{eq:loss} and \cref{eq:advantage}.

\begin{table}
	\centering
	\caption{Maximum mean cumulative training reward, averaged over 5 runs and 100 episodes and state of the art by~\cite{terry2020revisiting}. The number in the parenthesis denotes the episode in which it was achieved.\label{Tab:comparison}.}
	\begin{tabular}{m{5cm}m{2.4cm}m{2.4cm}m{2.4cm}}
		\toprule
		&  Multiwalker           & Waterworld              & Pursuit                 \\ \midrule
		{V-RACER} (LDI, single policy) &  \textbf{19.80 (3.2k)} & \textbf{194.57 (19.2k)} & \textbf{124.22 (20.8k)} \\
	    PPO, ApeX DDPG, ApeX DQN &  13.67 (9.2k)          & 14.8 (6.3k)             & 77.63 (24k)             \\
		QMIX                      & -5.65 (17.8k)         & 1.62 (45.3k)            & 45.41 (59.6k)           \\
		MADDPG                  &  -33.95(44.2k)         & -1.81 (41k)             & 4.04 (28.8k)            \\ \bottomrule
	\end{tabular}
\end{table}

In \cref{fig:SISL results} we show the results for the three SISL environments. We find that the LDI algorithm results are  superior to those of the other algorithmic variants. More specifically, while in the Multiwalker (a,d) environment the improvement is minimal, the advantage is clearly noticeable  in the Waterworld (b,e) and Pursuit (c,f) environments.
We argue that the FDI and LDI algorithms dominate their cooperative counterparts (FDCo and LDCo) because of the credit assignment problem: After averaging the rewards it is difficult to identify which actions contribute most to the success of the agents.
Furthermore, sharing the experiences among all agents in a single policy is sufficient to resolve the non-stationary issue and using \cref{eq:dependent iw} does not produce valuable information. In contrast, we find that taking the product of the individual importance weights harms the update. 
The relative order in terms of  performance for the employed algorithms is similar when training multiple policies. Interestingly, the maximal performance of the LDI, reached during training, is worse for multiple policies. During the evaluation, however, the single policy variant is outperformed (see \cref{tab:evaluation}). Another important difference arises for the LDCo. For multiple policies experience sharing does not remove the non-stationary issue. Together with the credit assignment problem, it hinders the convergence of the LDCo method. The FDCo alleviates this problem. By adding the information of the other agents in the importance weight (\cref{eq:dependent iw}), the destructive effect of non-stationary and the credit assignment problem is resolved.

Additionally, we compare our results when training a single policy with the results in~\citet{terry2020revisiting}, which, to the best of our knowledge, represent the state of the art on the SISL environments. In \Cref{Tab:comparison} we show a comparison between the  best results in \citet{terry2020revisiting} and our best performing alternative (LDI). We find that on all environments LDI outperforms the existing results by a margin.

In Multiwalker, we find that the LDI is the preferred algorithm. This can be especially seen during the first 8k episodes. After that, the returns for some runs drop significantly and  yield a lower asymptotic median return. This behaviour is often observed in Walker-like environments. In the failing runs the agents start falling while trying to move faster and faster. In order to avoid this catastrophic failure the algorithm converges to a safe, but sub-optimal policy. 
Nonetheless, the models that consider the local dynamics are the better performing alternatives. 
In the Multiwalker the LDI obtains a final return of 4.09, which is comparable to the attained return with the LDCo (3.73). The other variants do not achieve positive rewards (-31.68 for FDI and -38.11 for FDCo). 
We observe that in LDI certain runs identify a policy which consistently achieves a cumulative median return of around $20$, however the algorithm does not  learn the associated policy in every run.
Here, the state of the art~\cite{terry2020revisiting} achieves the highest cumulative reward when using a multi agent version of PPO~\cite{Schulman2017}. From \cref{Tab:comparison} it can be seen, that the LDI outperforms the multi agent version of PPO.

In the second continuous action environment Waterworld, we see clear differences between the algorithms. The LDI outperforms the other variants significantly. We observe, that the FDI and the LDCo show similar performance.
In Waterworld, it seems that correlating gradient updates (FDI) and rewards (LDCo) have similar negative effects. This effects seem to sum up when considering both correlations (FDCo). 
In Waterworld the state of the art is a MA version of ApeX DDPG~\cite{Horgan2018}. The best variant of the proposed algorithm outperforms the existing results by 13X. 

Finally, we discuss the results on the Pursuit benchmark. The negative effect arising from the credit assignment problem (LDCo) is larger than when correlating the update (FDI), which is contrary to our findings in Waterworld. Combining both assumptions (FDCo) still is the worst performing alternative. As can be seen in \cref{Tab:comparison}, the present best performing alternative outperforms the state of the art (a MA version of ApeX DQN~\cite{Horgan2018}). Comparing the maximal achieved mean returns shows that ReF-ER MARL provides 1.6X  higher return. 

\begin{table}
\centering
\caption{Testing the optimal policy for LDI on the SISL environments. The mean, maximal, and minimal cumulative reward is computed over 50 episodes. We highlight the better result.\label{tab:evaluation}}
\begin{tabular}{m{2.7cm}m{1cm}m{0.75cm}m{0.75cm}m{0.75cm}m{0.75cm}m{0.75cm}m{0.75cm}m{0.75cm}m{0.75cm}}
\toprule
                         &   \multicolumn{3}{c}{Multiwalker}            & \multicolumn{3}{c}{Waterworld}                   & \multicolumn{3}{c}{Pursuit}                      \\
                            \cmidrule(r){2-4} \cmidrule(r){5-7}  \cmidrule(r){8-10}
                                &  Mean      & Max      & Min       & Mean        & Max        & Min         & Mean        & Max        & Min         \\ \midrule
Single Policy                   &  \textbf{1.0} & \textbf{1.9}          & \textbf{0.4} & 202.3          & 271.8          & 140.6          & 56.8           & 84.7           & 11.6           \\
Multiple Policies &  -0.1         & 1.5          & -2.2         & \textbf{236.4}          & \textbf{303.4} & \textbf{191.0} & \textbf{130.3} & \textbf{146.6} & \textbf{112.8} \\ \bottomrule
\end{tabular}%
\end{table}

\section{ReF-ER and Scientific Multi-Agent Reinforcement Learning}

\begin{figure*}
	\centering
	\includegraphics[width=\linewidth,trim=100 170 100 150, clip]{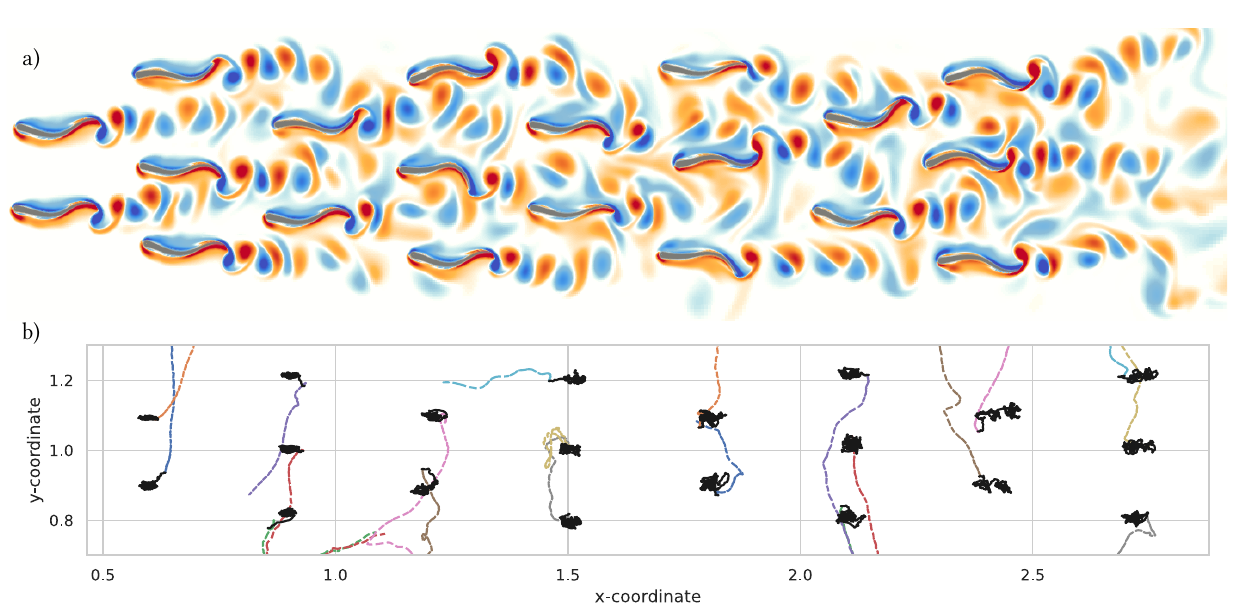}
	\caption{Structure of the school for 100 swimming periods. Figure (a) shows the swimmers in gray, where the colors visualize the vorticity field (blue negative and red positive). Figure (b) displays the trajectories of the controlled (black) and uncontrolled (dashed, color) swimmers relative to the center of mass of the school. An animation of both schools can be found in the appendix.}\label{fig:swimmers}
\end{figure*}

We test LDI with multiple policies in the problem of fish schooling in the presence of strong hydrodynamic interactions, in high fidelity simulations of the Navier-Stokes equations~\cite{Verma2018}.
We consider 20 self-propelled swimmers (figure \cref{fig:swimmers}) interacting  through their vortex wakes. The non-linear vortex interactions make this a challenging control problem ~\cite{Huiyang2022}.
Here, each swimmer observes a 16-dimensional state, encoding the ability to sense the environments via sight, flow sensing, and proprioception. 
Given the observation of the environment, the swimmers can control their motion via two actions that allow steering and changing the swimming velocity. Their reward is composed of the instantaneous swimming efficiency and a penalty term for collisions.
In \cref{fig:swimmers} we show the trajectories relative to the center of mass of the school. Controlled schooling behaviour is observed for 100 tail beat  periods whereas without control the swimmers collide at 1/5th of this time. Control for high fidelity simulations has a  high computational cost, which requires asynchronous training and data collection. We use one compute node to train the neural networks and 64 independent simulations with two nodes for data collection. In total, the run took four days on 129 nodes of a Cray XC50 system equipped with a 12 core Intel® Xeon® CPU and one NVIDIA® Tesla® P100 GPU. This suggests the suitability of our method to expensive  scientific computing  applications.

\section{Discussion}

We present ReF-ER MARL, the multi-agent generalization of the V-RACER algorithm with ReF-ER~\cite{Novati2019}. The combination of V-RACER with ReF-ER has shown significant promise in several benchmark problems and challenging fluid dynamics applications. 

In the proposed  ReF-ER MARL the actions of the agents are independent while the probability distribution of an action is conditioned on the respective agent's state. We examine the effects of  different relations between the agents and vary the strength of their interactions.  This is achieved by modifying the value estimator and the importance weight. 
For the value estimator we distinguish an individual and a cooperative setting by either using the individual reward or the average of the rewards from all agents. The strength of the agents' interaction is controlled via the importance weight.  

We benchmark ReF-ER MARL on the Stanford Intelligent Systems Laboratory (SISL) Environments. We   compare the  value-estimates, and importance weights for different assumptions. 
In these collaborative environments one may expect that assuming a strong interaction and cooperative estimates for the state-value is  beneficial. However, we find that the preferred approach is to estimate the value using individual rewards, and to consider a local dynamics model.  We find that ReF-ER MARL using a single feed forward neural network outperforms the state of the art algorithms, that often relies on complex network architectures.
Finally, we test the proposed algorithms when training multiple policies. The median returns are lower during training. During testing, one controller per agent outperforms a shared controller. Besides robustness, we show that introducing stricter dependencies between the agents via the full dynamics (FDCo) cures the deadly combination of non-stationarity and reward averaging.
Furthermore, the LDI with multiple policies is applied to enforce coordinated swimming on  20 swimmers experiencing   hydrodynamic interactions in high-fidelity simulations of the Navier-Stokes equation. It enables stable schooling formations for up to 100 tail-beat periods( the formation brakes up after 20 tail beats without control).

\section*{Broader Impact}

Multiagent reinforcement learning has been recently extended  to solve accurately underresolved partial differential equations  by casting the closure problem  as control with continuous  action spaces~\cite{Novati2021,Bae2022}. The success of this effort relied on using  V-RACER with ReF-ER~\cite{Novati2019}.  We extend this framework  to a variety of multiagent models. We benchmark the proposed algorithms in the SISL environments as well as a computationally expensive application of fish schooling  involving direct numerical simulations of the Navier-Stokes equation. The present work can serve as  a reference  for  scientific reinforcement learning community, in applications that require continuous control.

This research doesn't put anyone at disadvantage nor does a failure of the proposed method yield unfavorable consequences. This work does not leverage any biases in the data.



\begin{ack}
We acknowledge the infrastructure and support of CSCS, providing the necessary computational resources under project s929.
\end{ack}


\input{supplementary-material.tex}

\end{document}

%% file: macros.tex
\newcommand{\st}{s}
\newcommand{\stS}{{\cal S}}

\newcommand{\ac}{a}
\newcommand{\acS}{{\cal A}}

\newcommand{\re}{r}

\newcommand{\sti}{s^{(i)}}

\newcommand{\vstate}{\boldsymbol{\st}}
\newcommand{\vaction}{\boldsymbol{\ac}}
\newcommand{\vreward}{\boldsymbol{\re}}

\newcommand{\COMMA}{\,,}
\newcommand{\PERIOD}{\,.}

%% file: supplementary-material.tex
\appendix

\section{V-RACER for discrete action space}
The V-RACER paper \cite{Novati2019} presents the continuous action version of V-RACER. In this work, we extend V-RACER for the discrete action domain. For discrete action environments we employ a neural network which takes as an input the state and outputs the state-value estimate, the energies of each action $\epsilon_i$ for $i=1,\dots,|\mathcal{A}|$ and an inverse temperature $\beta$ parameter. The probability $p_i$ of sampling action $a_i$ is calculated according to the Boltzmann distribution 
\begin{equation}
    p_i = \frac{\exp{(-\epsilon_i) \beta}}{\sum_{k=1}^{|\mathcal{A}|}\exp{(-\epsilon_k) \beta}}\PERIOD
\end{equation}
Here we omit the derivation of the importance weight and the gradient thereof, as well as the gradient of the Kullback-Leibler divergence of the current policy from the past policy. 

\section{Clipped Normal Distribution}\label{sec:clipped normal}
We use the clipped normal distribution to enforce actions within an interval $[a,b]\subset{\mathbb{R}}$. The probability density function of the clipped normal distribution~\cite{Fujita2018} is given by 
\begin{equation}\label{eq:clipped_gaussian}
\begin{split}
    f(x;\mu,\sigma)&=\delta(x-a)F_{\cal N}(a;\mu,\sigma)+\mathbb{1}_{a<x<b}f_{\cal N}(x;\mu,\sigma)+\delta(x-b)[1-F_{\cal N}(b;\mu,\sigma)]\COMMA
\end{split}
\end{equation}
where  $\mathbb{1}_{a\leq x<b}$ is the indicator function that is $1$ inside $[a,b]$ and $0$ outside, $\delta$ denotes the Dirac delta distribution, and $f_{\cal N}(x;\mu,\sigma)$ is the density function of the normal distribution with mean $\mu$ and standard deviation $\sigma$. In contrast to the squashed normal distribution \cite{haarnoja2018}, the clipped normal distribution retains higher probability mass towards the action bounds. We found that clipping performs superior than applying squashing. In the following we derive the gradient of the Kullback-Leibler divergence, and the importance weight and the gradient thereof.

\subsection{Kullback-Leibler Divergence}

The definition of the Kullback-Leibler divergence is given by
\begin{equation}
D_{\mathrm{KL}}(p \| q)=\int_{-\infty}^{\infty}  \log \left[\frac{p(x;\mu_p,\sigma_p)}{q(x;\mu_q,\sigma_q)}\right] p(x;\mu_p,\sigma_p)\mathrm{d} x\,\PERIOD
\end{equation}
Plugging in the expression from \cref{eq:clipped_gaussian} we find
\begin{equation}
\begin{split}
    D_{\mathrm{KL}}(p \| q)&=\log\left[\frac{F_{\cal N}(a;\mu_p,\sigma_p)}{F_{\cal N}(a;\mu_q,\sigma_q)}\right]F_{\cal N}(a;\mu_p,\sigma_p)\\
    &+\underbrace{\int\limits_a^b\log\left[\frac{f_{\cal N}(x;\mu_p,\sigma_p)}{f_{\cal N}(x;\mu_q,\sigma_q)}\right]f_{\cal N}(x;\mu_p,\sigma_p)\mathrm{d} x}_{I}\\
    &+\log\left[\frac{1-F_{\cal N}(b;\mu_p,\sigma_p)}{1-F_{\cal N}(b;\mu_q,\sigma_q)}\right]\left[1-F_{\cal N}(b;\mu_p,\sigma_p)\right]\PERIOD
\end{split}
\end{equation}
Plugging in the expression for the normal distribution we can write the integral expression for $x\in(a,b)$ as
\begin{equation}
I=\frac{1}{\sqrt{2\pi}\sigma_p}\int_{a}^{b} \left\{\log\left(\frac{\sigma_q}{\sigma_p}\right)-\frac{1}{2}\left[\frac{(x-\mu_p)^2}{\sigma_p^2}-\frac{(x-\mu_q)^2}{\sigma_q^2}\right]\right\}\exp\left[-\frac{(x-\mu_p)^2}{2\sigma_p^2}\right]\mathrm{d} x\PERIOD
\end{equation}
Using a substitution of variables $x'=\frac{x-\mu_p}{\sqrt{2}\sigma_p}$, the integral reads
\begin{equation}
\begin{split}
I&=\frac{1}{\sqrt{\pi}}\int_{\frac{a-\mu_p}{\sqrt{2}\sigma_p}}^{\frac{b-\mu_p}{\sqrt{2}\sigma_p}} \underbrace{\left[\log\left(\frac{\sigma_q}{\sigma_p}\right)-x'^2+\frac{1}{2}\frac{(\sqrt{2}\sigma_px'+\mu_p-\mu_q)^2}{\sigma_q^2}\right]}_{Q}e^{-x'^2}\mathrm{d} x'\PERIOD
\end{split}
\end{equation}
We expand $Q$ giving
\begin{equation}
\begin{split}
Q&=\log\left(\frac{\sigma_q}{\sigma_p}\right)-x'^2-\frac{2\sigma_p^2x'^2+2\sqrt{2}\sigma_p(\mu_p-\mu_q)x'+(\mu_p-\mu_q)^2}{2\sigma_q^2}\\
&=\underbrace{\log\left(\frac{\sigma_q}{\sigma_p}\right)+\frac{(\mu_p-\mu_q)^2}{2\sigma_q^2}}_{C_1}+\underbrace{\sqrt{2}\sigma_p\frac{(\mu_p-\mu_q)}{\sigma_q^2}}_{C_2}x'-\underbrace{\left(1-\frac{\sigma_p^2}{\sigma_q^2}\right)}_{C_3}x'^2\PERIOD\\
\end{split}
\end{equation}
Using the identities 
\begin{equation}\label{eq:identities}
\begin{split}
\int_{a}^{b} e^{-x^{2}} \mathrm{d} x&=\frac{\sqrt{\pi}}{2}\left[\operatorname{erf}\left(b\right)-\operatorname{erf}\left(a\right)\right]\COMMA\\
\int_{a}^{b} xe^{-x^{2}} \mathrm{d} x&=-\frac{1}{2} \left[\exp\left(b^{2}\right)-\exp\left(a^{2}\right)\right]\COMMA\\
%
\int_{a}^{b} x^2e^{-x^{2}} \mathrm{d} x&=\frac{\sqrt{\pi}}{4}\left[\operatorname{erf}(b)-\operatorname{erf}(a)\right]-\frac{1}{2}\left[b \exp(-b^2) -a \exp(-a^2)\right]\,,
\end{split}
\end{equation}
the integration can be performed
\begin{equation}
\begin{split}
    D_{KL}(p||q)&=\log\left[\frac{F_{\cal N}(a;\mu_p,\sigma_p)}{F_{\cal N}(a;\mu_q,\sigma_q)}\right]F_{\cal N}(a;\mu_p,\sigma_p)\\
    &+\frac{1}{2}\left[\log\left(\frac{\sigma_q}{\sigma_p}\right)+\frac{(\mu_p-\mu_q)^2}{2\sigma_q^2}\right]\left[\operatorname{erf}\left(\frac{b-\mu_p}{\sigma_p \sqrt{2}}\right)-\operatorname{erf}\left(\frac{a-\mu_p}{\sigma_p \sqrt{2}}\right)\right]\\
    &-\frac{1}{\sqrt{2\pi}}\sigma_p\frac{(\mu_p-\mu_q)}{\sigma_q^2}\left\{ \exp\left[-\frac{(b-\mu_p)^{2}}{2\sigma_p^2}\right]-\exp\left[-\frac{(a-\mu_p)^{2}}{2\sigma_p^2}\right]\right\}\\
    &-\frac{1}{4}\left(1-\frac{\sigma_p^2}{\sigma_q^2}\right) \left[\operatorname{erf}\left(\frac{b-\mu_p}{\sqrt{2}\sigma_p}\right)-\operatorname{erf}\left(\frac{a-\mu_p}{\sqrt{2}\sigma_p}\right)\right]\\
    &+\frac{1}{2}\frac{1}{\sqrt{\pi}}\left(1-\frac{\sigma_p^2}{\sigma_q^2}\right)\left[\left(\frac{b-\mu_p}{\sqrt{2}\sigma_p}\right)\exp\left(-\frac{(b-\mu_p)^{2}}{2\sigma_p^2}\right)-\left(\frac{a-\mu_p}{\sqrt{2}\sigma_p}\right)\exp\left(-\frac{(a-\mu_p)^{2}}{2\sigma_p^2}\right)\right]\\
    &+\log\left[\frac{1-F_{\cal N}(b;\mu_p,\sigma_p)}{1-F_{\cal N}(b;\mu_q,\sigma_q)}\right]\left[1-F_{\cal N}(b;\mu_p,\sigma_p)\right]\\
    &=\log\left[\frac{F_{\cal N}(a;\mu_p,\sigma_p)}{F_{\cal N}(a;\mu_q,\sigma_q)}\right]F_{\cal N}(a;\mu_p,\sigma_p)\\
    &+\frac{1}{2}\left[\log\left(\frac{\sigma_q}{\sigma_p}\right)+\frac{(\mu_p-\mu_q)^2}{2\sigma_q^2}-\frac{1}{2}\left(1-\frac{\sigma_p^2}{\sigma_q^2}\right)\right]\left[\operatorname{erf}\left(\frac{b-\mu_p}{\sigma_p\sqrt{2}}\right)-\operatorname{erf}\left(\frac{a-\mu_p}{\sigma_p \sqrt{2}}\right)\right]\\
    &+\frac{1}{\sqrt{2\pi}}\left[\frac{1}{2}\left(1-\frac{\sigma_p^2}{\sigma_q^2}\right)\left(\frac{b-\mu_p}{\sigma_p}\right)-\frac{\sigma_p(\mu_p-\mu_q)}{\sigma_q^2}\right]\exp\left(-\frac{(b-\mu_p)^{2}}{2\sigma_p^2}\right)\\
    &-\frac{1}{\sqrt{2\pi}}\left[\frac{1}{2}\left(1-\frac{\sigma_p^2}{\sigma_q^2}\right)\left(\frac{a-\mu_p}{\sigma_p}\right)-\frac{\sigma_p(\mu_p-\mu_q)}{\sigma_q^2}\right]\exp\left(-\frac{(a-\mu_p)^{2}}{2\sigma_p^2}\right)\\
    &+\log\left[\frac{1-F_{\cal N}(b;\mu_p,\sigma_p)}{1-F_{\cal N}(b;\mu_q,\sigma_q)}\right]\left[1-F_{\cal N}(b;\mu_p,\sigma_p)\right]\;.
\end{split}
\end{equation}
Using the derivatives of the cumulative distribution function with respect to the mean and standard deviation
\begin{equation}\label{eq:Cumulative_Gaussian_derivative}
    \begin{split}
    \frac{F_{\cal N}(x;\mu,\sigma)}{\partial \mu_q}&=-f_{\cal N}(x;\mu,\sigma)\,,\\
    \frac{F_{\cal N}(x;\mu,\sigma)}{\partial \sigma_q}&=-\frac{x-\mu}{\sigma}f_{\cal N}(x;\mu,\sigma)\COMMA
    \end{split}
\end{equation}
and the derivative of the error function
\begin{equation}\label{eq:derivative_errorfunction}
    \frac{\partial}{\partial z} \operatorname{erf}(z)=\frac{2}{\sqrt{\pi}} e^{-z^{2}}\COMMA
\end{equation}
the derivative of the Kullback-Leibler divergence with respect to $\mu_q$ and $\sigma_q$ is given by
\begin{equation}
\begin{split}
    \frac{\partial D_{\mathrm{KL}}(p \| q)}{\partial \mu_q}&=\frac{F_{\cal N}(a;\mu_p,\sigma_p)}{F_{\cal N}(a;\mu_q,\sigma_q)}f_{\cal N}(a;\mu_q,\sigma_q)\\
    &-\frac{1}{2}\frac{\mu_p-\mu_q}{\sigma_q^2}\left[\operatorname{erf}\left(\frac{b-\mu_p}{\sqrt{2}\sigma_p}\right)-\operatorname{erf}\left(\frac{a-\mu_p}{\sqrt{2}\sigma_p}\right)\right]\\
    &+\frac{1}{\sqrt{2\pi}}\frac{\sigma_p}{\sigma_q^2}\left[\exp\left(-\frac{(b-\mu_p)^{2}}{2\sigma_p^2}\right)-\exp\left(-\frac{(a-\mu_p)^{2}}{2\sigma_p^2}\right)\right]\\
    &-\frac{1-F_{\cal N}(b;\mu_p,\sigma_p)}{1-F_{\cal N}(b;\mu_q,\sigma_q)}f_{\cal N}(b;\mu_q,\sigma_q)\\
    \frac{\partial D_{\mathrm{KL}}(p \| q)}{\partial \sigma_q}&=
    \frac{a-\mu_q}{\sigma_q}
    \frac{F_{\cal N}(a;\mu_p,\sigma_p)}{F_{\cal N}(a;\mu_q,\sigma_q)}f_{\cal N}(a;\mu_q,\sigma_q)\\
    &+\frac{1}{2}\left[\frac{1}{\sigma_q}-\frac{(\mu_p-\mu_q)^2}{\sigma_q^3}-\frac{\sigma_p^2}{\sigma_q^3}\right]\left[\operatorname{erf}\left(\frac{b-\mu_p}{\sigma_p \sqrt{2}}\right)-\operatorname{erf}\left(\frac{a-\mu_p}{\sigma_p \sqrt{2}}\right)\right]\\
    &+\frac{1}{\sqrt{2\pi}}\left[\frac{\sigma_p^2}{\sigma_q^3}\left(\frac{b-\mu_p}{\sigma_p}\right)+\frac{2\sigma_p(\mu_p-\mu_q)}{\sigma_q^3}\right]\exp\left(-\frac{(b-\mu_p)^{2}}{2\sigma_p^2}\right)\\
    &-\frac{1}{\sqrt{2\pi}}\left[\frac{\sigma_p^2}{\sigma_q^3}\left(\frac{a-\mu_p}{\sigma_p}\right)+\frac{2\sigma_p(\mu_p-\mu_q)}{\sigma_q^3}\right]\exp\left(-\frac{(a-\mu_p)^{2}}{2\sigma_p^2}\right)\\
    &-\frac{b-\mu_q}{\sigma_q}\frac{1-F_{\cal N}(b;\mu_p,\sigma_p)}{1-F_{\cal N}(b;\mu_q,\sigma_q)}f_{\cal N}(b;\mu_q,\sigma_q)\PERIOD
\end{split}
\end{equation}

\subsection{Importance Weight}
The gradient of the importance weight
\begin{equation}
    \operatorname{IW}(x; \mu_q, \sigma_q, \mu_p, \sigma_p) =\begin{cases} \frac{F_{\cal N}(a;\mu_q,\sigma_q)}{F_{\cal N}(a;\mu_p,\sigma_p)}\,, & \text{if }x=a\,,\\
    \frac{f_{\cal N}(x;\mu_q,\sigma_q)}{f_{\cal N}(x;\mu_p,\sigma_p)}\,, & \text{if }a<x<b\,, \\
     \frac{1-F_{\cal N}(b;\mu_q,\sigma_q)}{1-F_{\cal N}(b;\mu_p,\sigma_p)}\,, & \text{if }x=b\,,
    \end{cases}
\end{equation}
with respect to the parameters $\mu_q$ and $\sigma_q$ can be computed using 
\begin{equation}\label{eq:Gaussian_derivative}
    \begin{split}
    \frac{\partial f_{\cal N}(x;\mu,\sigma)}{\partial \mu_q}&=\frac{x-\mu}{\sigma^2}f_{\cal N}(x;\mu,\sigma)\;,\\
    \frac{\partial f_{\cal N}(x;\mu,\sigma)}{\partial \sigma_q}&=\Big(-\frac{1}{\sigma}+\frac{(x-\mu)^2}{\sigma^3}\Big)f_{\cal N}(x;\mu,\sigma)=\frac{\mu^2-\sigma^2-2\mu x + x^2}{\sigma^3}f_{\cal N}(x;\mu,\sigma)\;,
    \end{split}
\end{equation}
and \cref{eq:Cumulative_Gaussian_derivative} as
\begin{equation}
\frac{\partial \operatorname{IW}}{\partial \mu_q}=\begin{cases} 
        \frac{-f_\mathcal{N}(a;\mu_q,\sigma_q)}{F_{\cal N}(a;\mu_p,\sigma_p)}\,, & \text{if }x=a\,,\\
    \frac{(x-\mu_q)f_\mathcal{N}(x;\mu_q,\sigma_q)}{\sigma_q^2f_{\cal N}(x;\mu_p,\sigma_p)}\,, & \text{if }a<x<b\,, \\
     \frac{f_\mathcal{N}(b;\mu_q,\sigma_q)}{1-F_{\cal N}(b;\mu_p,\sigma_p)}\,, & \text{if }x=b\,.
    \end{cases}
\end{equation}
\begin{equation}
\frac{\partial \operatorname{IW}}{\partial \sigma_q}=\begin{cases} 
    -\frac{a-\mu_q}{\sigma_q}\frac{f_\mathcal{N}(a;\mu_q,\sigma_q)}{F_{\cal N}(a;\mu_p,\sigma_p)}\,, & \text{if }x=a\,,\\
    \frac{((x-\mu_q)^2-\sigma_q^2)f_{\cal N}(x;\mu_q,\sigma_q)}{\sigma_q^3f_{\cal N}(x;\mu_p,\sigma_p)}\,, & \text{if }a<x<b\,, \\
     \frac{b-\mu_q}{\sigma_q}\frac{f_\mathcal{N}(b;\mu_q,\sigma_q)}{1-F_{\cal N}(b;\mu_p,\sigma_p)}\,, & \text{if }x=b\,.
    \end{cases}
\end{equation}

\newpage
\section{Algorithms}\label{sec:Algos}

The general MARL loop is presented in \cref{algo:MARL}.
\SetAlgoNoLine
\begin{algorithm}[h]
\caption{Multi-Agent Reinforcement Learning (Synchronous Variant)}\label{algo:MARL}
    \SetKwInOut{Input}{Input}
    \Input{Environment function $D(\boldsymbol{s},\boldsymbol{a})$, Replay Memory ${\cal RM}$, Neural Network(s) $NN$ for agents $i=1,\dots,N$, Termination Criteria ${\cal T}$}
    \BlankLine
    \While{not ${\cal T}$}{
        \BlankLine
        \textsc{// Collecting Experiences}
        \BlankLine
        Sample Initial States ${s}^{(i)}_0$ for $i=1,\dots,N$
        \BlankLine
        \For{$t\in 1,\dots, T$}{
            Forward Neural Network $V^{\pi}({s}^{(i)}_{t-1}),\pi_{t-1}(\cdot|{s}^{(i)}_{t-1})=NN({s}^{(i)}_{t-1};\boldsymbol{\vartheta},\boldsymbol{\omega})$ for $i=1,\dots,N$
            \BlankLine
            Sample Actions ${a}^{(i)}_{t-1}\sim \pi_{t-1}(\cdot|{s}^{(i)}_{t-1})$ for $i=1,\dots,N$
            \BlankLine
            Run Environment Function $\boldsymbol{r}_{t-1}, \boldsymbol{s}_{t}=D(\boldsymbol{a}_{t-1}, \boldsymbol{s}_{t-1})$
        }
    \textsc{// Postprocess Episode}
    \BlankLine
    Save Episode ${\cal E}_k=\{(\boldsymbol{s}_t, \boldsymbol{a}_t, \boldsymbol{r}_t, \boldsymbol{V}_t,\boldsymbol{\pi}_t)\}_{t=0}^{T_k}$ in Replay Memory ${\cal RM}$
    \BlankLine
    Set $\hat{V}_{T}^{\mathrm{tbc},(i)}=V^{\pi}({s}_T^{(i)})$ for $i=1,\dots,N$
    \BlankLine
    \For{$t\in T-1,\dots, 1$}{
        Compute $\hat V_t^{\mathrm{tbc},(i)}=V^{\pi}({s}_t^{(i)})+\bar{\rho}_{t}\left[f(\boldsymbol{r}_{t})+\gamma \hat{V}_{t+1}^{\mathrm{tbc},(i)}-V^{\pi}({s}_t^{(i)})\right]$ for $i=1,\dots,N$
        \BlankLine
        Add $\hat{V}^{\mathrm{tbc},(i)}_t$ for $i=1,\dots,N$ to Replay Memory ${\cal RM}$
    }
    
    \BlankLine
    \textsc{// Training Neural Network}
    \BlankLine
    \For{$i\in 1,\dots,T$}{
        miniBatch = generateMinibatch()
        
        trainPolicy( miniBatch )
        
        updateReFERparams()
    }
  }
\end{algorithm}\ \\
Here $f$ implements the chosen relation between the agents. The function \textsc{updateReFERparams} updates the ReF-ER parameters, and \textsc{generateMinibatch} samples a minibatch of experiences from the replay memory. \textsc{trainPolicy} implements the learning algorithm and is detailed in \cref{algo:trainPolicyGeneral}. Depending on the chosen variant of the algorithm the function and \textsc{computeImportanceWeight} and \textsc{computePolicyGradient} implement the respective variant as described in the main text. The function \textsc{isOnPolicy} classifies the experience as on- or off-policy.
\SetAlgoNoLine
\begin{algorithm}[!htbp]
\captionof{algorithm}{\textsc{trainPolicy}}\label{algo:trainPolicyGeneral}
    \SetKwInOut{Input}{Input}
    \Input{ miniBatch }
    \BlankLine
    $\hat{\boldsymbol{g}}^V(\boldsymbol{\vartheta})=\hat{\boldsymbol{g}}^{\pi}(\boldsymbol{\omega}) =\boldsymbol{0}$
    \BlankLine
    \For{ $(\boldsymbol{s}_b, \boldsymbol{a}_b, \boldsymbol{r}_b, \boldsymbol{\pi}_b)\in {miniBatch}$ }{
        \BlankLine
        \textsc{// Forward Neural Network}
        \BlankLine
        $V_{\boldsymbol{\vartheta}}^\pi(s_b^{(i)}), \pi_{\boldsymbol{\omega}}(s_{b}^{(i)})=NN({s}_b^{(i)};\boldsymbol{\vartheta},\boldsymbol{\omega})$ for $i=1,\dots,N$
        \BlankLine
        \textsc{// Backwards Update of Value Estimator}
        \BlankLine
        $b_{\text{first}}\leftarrow$ find first experience for episode containing $b$
        \BlankLine
        \For{$t\in b,\dots,b_{\text{first}}$}{
            $V_{f}^\pi(s_t^{(i)})\leftarrow$ scalarize value from replay memory
            \BlankLine
            $\hat V_t^{\mathrm{tbc},(i)}=V_f^{\pi}(s_{t}^{(i)})+\bar{\rho}_{t}\left[f(\boldsymbol{r}_{t})+\gamma \hat{V}_{t+1}^{\mathrm{tbc},(i)}-V_f^{\pi}(s_{t}^{(i)})\right]$ for $i=1,\dots,N$
        }
        \BlankLine
        \textsc{// Compute Importance Weight}
        \BlankLine
        importanceWeight = computeImportanceWeight($\boldsymbol{a}_b, \boldsymbol{\pi}_b, \boldsymbol{\pi}_{\boldsymbol{\vartheta}}$)
        \BlankLine
        \textsc{// Compute Value Gradient}
        \BlankLine
        $V_{f}^\pi(s_b^{(i)})\leftarrow$ scalarize value $V_{\boldsymbol{\vartheta}}^\pi(s_b^{(i)})$
        \BlankLine
        $\hat{\boldsymbol{g}}^{V,(i)}_{b}(\boldsymbol{\vartheta})=\frac{1}{N}\left[V_f^{\pi}(\sti_b)-\hat{V}_{b}^{\mathrm{tbc},(i)}\right]\nabla_{\boldsymbol{\vartheta}}V_{\boldsymbol{\omega}}^\pi(s_b^{(i)})$ for $i=1,\dots,N$
        \BlankLine
        \textsc{// Compute Policy Gradient}
        \BlankLine
        $\hat{\boldsymbol{g}}^{\text{KL},(i)}_{b}(\boldsymbol{\omega})=\nabla_{\boldsymbol{\omega}}D_{\mathrm{KL}}\left({\pi}_b\| \pi_{\boldsymbol{\omega}}\right)(s_{b}^{(i)})$ for $i=1,\dots,N$
        \BlankLine
        \eIf{ isOnPolicy(importanceWeight) }{
            $\hat{\boldsymbol{g}}^{(i)}_{b}(\boldsymbol{\omega})$ =computePolicyGradient(importanceWeight,$V_{\boldsymbol{\vartheta}}^\pi(s_b^{(i)}), \pi_{\boldsymbol{\omega}}(s_{b}^{(i)})$)
        }
        {
            $\hat{\boldsymbol{g}}_{b}^{(i)}(\boldsymbol{\omega})= \boldsymbol{0}$
        }
        \textsc{// Accumulate Gradient for Value Function}
        \BlankLine
        $\hat{\boldsymbol{g}}^{V}(\boldsymbol{\vartheta})=\hat{\boldsymbol{g}}^{V}(\boldsymbol{\vartheta})+ \frac{1}{N\cdot|miniBatch|}\hat{\boldsymbol{g}}^{V,(i)}_{b}(\boldsymbol{\vartheta})$ for $i=1,\dots,N$
        \BlankLine
        \textsc{// Accumulate Gradient for Policy}
        \BlankLine
        $\hat{\boldsymbol{g}}^{\pi}(\boldsymbol{\omega}) =\hat{\boldsymbol{g}}^{\pi}(\boldsymbol{\omega}) + \frac{1}{N\cdot|miniBatch|}(\beta \hat{\boldsymbol{g}}^{(i)}_b(\boldsymbol{\omega}) + (1 - \beta)\hat{\boldsymbol{g}}^{\text{KL},(i)}_{b}(\boldsymbol{\omega}))$ for $i=1,\dots,N$
    }
    \BlankLine
    \textsc{// Update Hyperparameters}
    \BlankLine
    $\boldsymbol{\vartheta}\leftarrow \boldsymbol{\vartheta}-\eta \hat{\boldsymbol{g}}^V(\boldsymbol{\vartheta})$
    \BlankLine
    $\boldsymbol{\omega}\leftarrow \boldsymbol{\omega}-\eta \hat{\boldsymbol{g}}^{\pi}(\boldsymbol{\omega})$
\end{algorithm}

\newpage
\bibliographystyle{acm}
\bibliography{bibliography}